\documentclass[10pt,twocolumn,letterpaper]{article}

\usepackage{cvpr}              % To produce the CAMERA-READY version
\usepackage{tikz}
\usepackage{cancel}
\usepackage{pifont}% http://ctan.org/pkg/pifont
\newcommand{\cmark}{\ding{51}}%
\newcommand{\xmark}{\ding{55}}%
\definecolor{cvprblue}{rgb}{0.21,0.49,0.74}
\usepackage[pagebackref,breaklinks,colorlinks,citecolor=cvprblue]{hyperref}

\def\paperID{11915} % *** Enter the Paper ID here
\def\confName{CVPR}
\def\confYear{2024}

\title{Prompt Augmentation for Self-supervised Text-guided Image Manipulation}

\author{Rumeysa Bodur$^1$,  Binod Bhattarai$^2$,  Tae-Kyun Kim$^{1,3}$ \\%\thanks{Thanks to XYZ agency for funding.}}\\
$^1$Imperial College London, UK,  $^2$University of Aberdeen, UK,  $^3$KAIST, South Korea \\
\tt\small{r.bodur18@imperial.ac.uk, binod.bhattarai@abdn.ac.uk, kimtaekyun@kaist.ac.kr}}
\begin{document}
\maketitle
\begin{abstract}
Text-guided image editing finds applications in various creative and practical fields. While recent studies in image generation have advanced the field, they often struggle with the dual challenges of coherent image transformation and context preservation. In response, our work introduces prompt augmentation, a method amplifying a single input prompt into several target prompts, strengthening textual context and enabling localised image editing. Specifically, we use the augmented prompts to delineate the intended manipulation area. We propose a Contrastive Loss tailored to driving effective image editing by displacing edited areas and drawing preserved regions closer. Acknowledging the continuous nature of image manipulations, we further refine our approach by incorporating the similarity concept, creating a Soft Contrastive Loss. The new losses are incorporated to the diffusion model, demonstrating improved or competitive image editing results on public datasets and generated images over state-of-the-art approaches.  
\end{abstract}    
\vspace{-0.6cm}
\section{Introduction}
\label{sec:intro}

\begin{figure}[t!]
  \centering
  \resizebox{0.82\linewidth}{!}{%
    \begin{tabular}{ccc}
      \includegraphics[width=3cm]{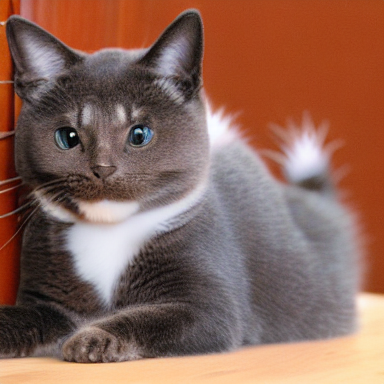} & \includegraphics[width=3cm]{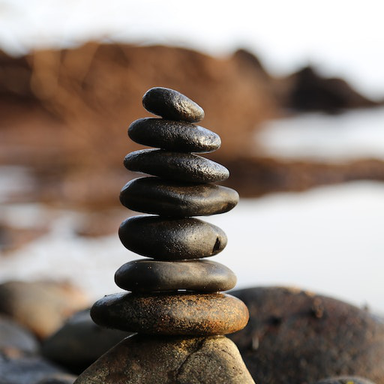} &
      \includegraphics[width=3cm]{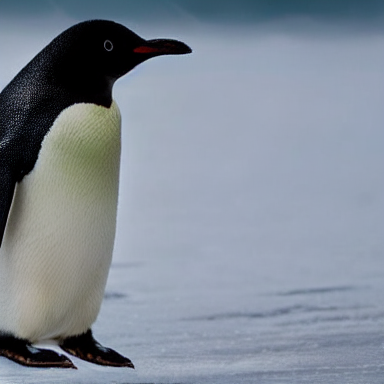} \\
      ``a \textcolor{brown}{leopard}'' &  ``a \textcolor{brown}{wedding} & ``a penguin \\ 
        & \textcolor{brown}{ cake}'' & \textcolor{brown}{walking on the beach''}\\

      \includegraphics[width=3cm]{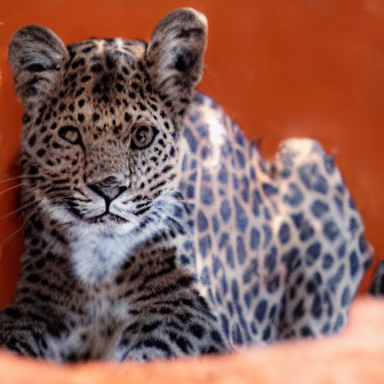}
      &\includegraphics[width=3cm]{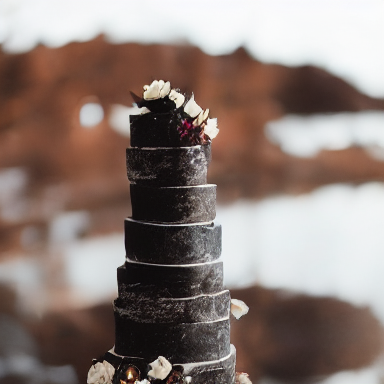}
      &\includegraphics[width=3cm]{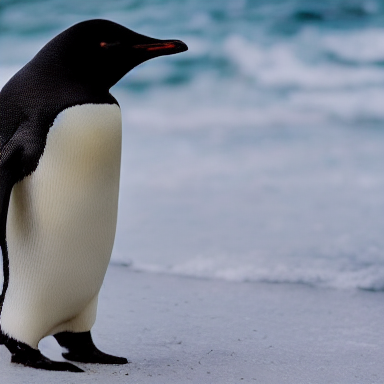}\\

    \end{tabular}%
  }
  \vspace{-0.2cm}

     \caption{\textbf{Text-guided image manipulation.} Illustrative examples generated by our method (bottom row) with localised manipulations based on given text prompts and input images (top row).}
\vspace{-0.5cm}

\label{fig:motivation}
\end{figure}

% taks decription and general challenges:
The field of text-guided image generation has made significant progress, particularly with the advent of diffusion models~\cite{rombach2022sd, singh2022highfidelity, ramesh2022dalle, kim2024arbitrary}, ushering in a new era of content creation. This progress extends to text-guided image manipulation, which offers a wide range of applications, spanning from artistic expression to the enhancement of image interpretability. Text-guided image manipulation involves altering an input image based on a user-provided textual prompt, such as changing the appearance or shape of objects, modifying the background, and adding, removing or replacing features. Some sample manipulations performed by our method can be seen in Figure~\ref{fig:motivation}. In this context, two intertwined challenges are present: the task of transforming image content in accordance with a provided textual description, and the need to preserve the salient aspects of the original visual information that remain contextually relevant. While recent methodologies \cite{meng2022sdedit, brooks2022instructpix2pix, hertz2022prompt} have excelled in the domain of image manipulation, they often encounter limitations in addressing both of these challenges concurrently.

% \begin{figure}
%     \centering
%     \begin{tikzpicture}
%         \draw[black, fill=lightgray] (0, 0) rectangle (6, 4); % Adjust the coordinates and size as needed
%     \end{tikzpicture}
%     \caption{\textbf{Text-guided image manipulation.} Illustrative examples demonstrating the effectiveness of our method in generating images with precise, localised manipulations based on given text prompts.}
%     \label{fig:placeholder}
% \end{figure}

%challenges with sota
Many pioneering text-guided image editing methods rely on text-guided image generation models~\cite{meng2022sdedit, hertz2022prompt, kawar2023imagic, diffedit, parmar2023zeroshoti2i, wu2023zeroshotediting}, which tend to be ill-equipped to address the nuanced challenges of image manipulation. Additionally, methods tailored specifically for image editing often require domain-specific training~\cite{bar2022text2live, kim2022diffusionclip}. These methods require training for each instance or translation domain, limiting their scalability across diverse tasks. Despite excelling within their trained domains, they often struggle with the broader generalisation needed for accommodating various editing scenarios. Although there exist generic methods, such as InstructPix2Pix~\cite{brooks2022instructpix2pix}, they often struggle to achieve both the faithful transformation of image content and the preservation of contextual details. 
Some existing methods incorporate masks at inference time~\cite{diffedit, avrahami2021blendeddiff, ramesh2022dalle} or during training~\cite{wang2023imageneditor}. Although these methods show a good performance at content preservation, Dall-E 2~\cite{ramesh2022dalle} and Imagen Editor~\cite{wang2023imageneditor} fall short as they treat it as an inpainting task without taking into consideration the masked content and DiffEdit~\cite{diffedit} shows weaknesses as it relies on a pre-trained image generation model that is not specifically trained for manipulation purposes and is sensitive to the mask detected at inference time. 

% prompt augmentaion and mask
Augmentation techniques have proven useful for enhancing image translation~\cite{cao2021remix, bodur2021densegeo, bodur2020hierarchical}, but their application in label space remains underexplored~\cite{bodur2020auglabel, lu2023specialistdiffusion, liu2023equitable, bhattarai2020inducing}.
Hence, to overcome the aforementioned challenges, this paper introduces the concept of prompt augmentation, which extends a single input prompt into a multitude of target prompts. This augmentation provides the model with a broader and more nuanced understanding of the image editing task at hand. Taking inspiration from DiffEdit~\cite{diffedit}, which uses user given source and target prompts to generate masks at inference time, we leverage augmented prompts to automatically derive an attention mask that serves as a guide for image editing at training time. This mask plays a pivotal role in distinguishing which regions of the input image should undergo alterations and which should remain unaltered, contributing to the task's overall precision in localising the desired edit.

% cl and soft-cl
In addition to facilitating attention map computation, augmented prompts empower us to establish relations between their respective manipulations. Specifically, areas falling into the obtained masked for each pair should exhibit dissimilarity, reflecting the diverse edits dictated by different prompts, while the unmasked regions should maintain similarity, preserving aspects unaffected by the prompt. With this aim we propose to incorporate a Contrastive Loss (CL) that serves this dual purpose, compelling the model to simultaneously displace edited areas while drawing closer the preserved regions. 
%By doing so, it reinforces the idea that the edited content aligns with the target prompt, thus facilitating the desired visual transformation while preserving the areas unaffected by the prompt. 
As manipulation is not a discreet task and the required amount of modification is influenced by the target prompt's relation to the source image, we propose to soften our contrastive loss by considering this relation. Building upon this foundation, we introduce a novel loss function, the Soft Contrastive Loss (Soft CL) to incorporate the concept of similarity between textual prompts. This approach results in a more dynamic and nuanced interaction between textual prompts and the image, contributing to a higher level of performance.

    \begin{itemize}
        \item We introduce prompt augmentation, expanding input prompts into multiple targets, in order to enhance contextual understanding for image editing, and compute a dynamic attention mask to guide editing localisation.
        \item We introduce CL to encourage effective editing, pulling preserved regions closer while pushing edited areas  to align with the target prompt. Our novel Soft CL incorporates similarity for dynamic prompt interaction, improving adaptability and performance.
        \item We assess the performance of our method through a comparative analysis with state-of-the-art approaches, achieving competitive results without relying on masks during inference or utilising a paired dataset.
    \end{itemize}

\vspace{-0.2cm}
\section{Related Work}
\label{sec:formatting}
\vspace{-0.2cm}
\noindent\textbf{Text-guided Image Manipulation.}
Text-guided image manipulation using diffusion models is an evolving field that harnesses diffusion models' progressive and attention-based architecture to facilitate fine-grained image adjustments. This research trajectory aims to enhance controllability and flexibility while maintaining the fidelity of the input image. 
% CLIP:
DiffusionCLIP~\cite{kim2022diffusionclip} leverages CLIP-based losses to guide the diffusion process where they fine-tune a pretrained diffusion model for a specific domain. 
% Specific domain or instance
Imagic~\cite{kawar2023imagic}, Text2Live~\cite{bar2022text2live} and Dreambooth~\cite{ruiz2022dreambooth} involve fine-tuning the entire model for each image, primarily generating variations for objects.
% cross-attention maps
Plug-and-Play~\cite{tumanyan2022plugandplay} explores the injection of spatial features and self-attention maps to maintain the overall structural integrity of the image. 
Prompt-to-Prompt (P2P)~\cite{hertz2022prompt} dispenses with fine-tuning, instead retaining image structure by assigning cross-attention maps from the original image to the edited one based on corresponding text tokens.
Another study~\cite{epstein2023selfguidance} uses self-guidance by constraining attention maps or intermediate activations to control the sampling process. Similarly, MasaCtrl~\cite{cao2023masactrl} proposes to exploit innate features to use mutual attention during inference.
% target images
InstructPix2Pix~\cite{brooks2022instructpix2pix} utilises P2P to generate target images by accommodating human-like instructions for image editing. 
% denoise-noise + masks
SDEdit~\cite{meng2022sdedit} adopts a two-step approach, where they first introduce noise into the input image and then employ the SDE prior for denoising, ultimately enhancing realism while aligning with user guidance.
BlendedDiffusion~\cite{avrahami2021blendeddiff} follows a similar approach but introduces manually created masks at inference time. Similarly, DiffEdit~\cite{diffedit} also uses masks at inference but automatically predicts them by using difference of latent noises obtained by input and target prompts provided by the user. 
Imagen Editor~\cite{wang2023imageneditor} incorporates masks during training by using an object detector for randomly masking out objects to propose inpainting masks during training.
In contrast, our method integrates masks during training, specifically tailored to the target manipulation. We determine these masks through the generation of multiple augmentations of the source prompt enabling us to employ a strategy akin to~\cite{diffedit} to obtain an attention mask.\\
\noindent\textbf{Prompt generation and augmentation for image generation.}
InstructPix-to-Pix~\cite{brooks2022instructpix2pix} proposes to generate instructions for image editing by using a GPT model that they train on manually constructed pairs of captions and instruction. 
Specialist Diffusion~\cite{lu2023specialistdiffusion} augments the prompts to define the same image with multiple captions that convey the \emph{same} meaning in order to improve the generalisation of the image generation network. They retrieve similar captions, replace words with synonyms or reflect the image augmentation, e.g. horizontal flip, in the text prompt. 
Another work~\cite{lin2023regeneration} proposes to generate random sentences including source and target domain in order to calculate a mean difference that will serve as a direction while editing. iEdit~\cite{bodur2023iedit} generates target prompts by changing words in the input caption in order to retrieve pseudo-target images and guide the model.  
In~\cite{liu2023equitable}, the authors propose to augment the prompts with cultural descriptions in order to reduce the culture bias in generative models. 
Our approach uniquely integrates prompt augmentation during training, automatically generating multiple augmentations of the source prompt that define possible target prompts. We leverage these augmented prompts to determine masks tailored specifically to the target manipulation, and to enable self-supervised learning, which enhance the model's ability to learn editing tasks.\\
\noindent\textbf{Contrastive Loss for image generation.} 
Being a powerful self-supersived learning method, contrastive learning ensures the consistency of image representations across various augmentations by comparing positive pairs against negative ones. While it has delivered remarkable outcomes in numerous fields, its application in image generation, and more specifically, image manipulation, remains relatively uncharted. In Cntr-GAN\cite{zhao2020cntrgan} the discriminator is trained with a loss to push different image representations apart, while drawing augmentations of the same image closer.
The first study to explore contrastive learning in the domain of image manipulation~\cite{park2020contrastive} proposes PatchNCE loss to minimise the distance between the feature representations of patches from a source image and corresponding patches from generated images while treating randomly sampled patches from other locations as negative samples. XMC-GAN~\cite{zhang2021cross} proposes to enforce text–image resemblance with a contrastive discriminator for text-to-image generation.\\
\noindent\textbf{Soft Contrastive Loss.} Most studies utilising contrastive learning rely on hard assignment of samples~\cite{duan2019deepeL, park2020contrastive,zhao2020cntrgan,zhang2021cross} while only a few explore soft assignment. In a study~\cite{kim2019deep} tested on image retrieving, the authors propose a log ratio loss to enforce the label and feature representation distances to be proportional. 
Another study~\cite{thoma2020softcl} follows a similar approach for visual localisation, in which they aim to have the ordering of euclidean distances between features respect the ordering of geometric proximity measure between the corresponding images.
Given the continuous nature of textual prompts and image manipulation, in our work, we leverage the similarity between augmented prompts to guide the soft assignment of image representations.
% \vspace{-0.2cm}
\section{Methodology}
% \vspace{-0.2cm}

% \begin{figure*}
%     \centering
%     \begin{tikzpicture}
%         \draw[black, fill=lightgray] (0, 0) rectangle (12, 4); % Adjust the coordinates and size as needed
%     \end{tikzpicture}
%     \caption{Overview of our method.}
%     \label{fig:method_overview}
% \end{figure*}

\begin{figure*}
    \centering
    \includegraphics[width=0.87\linewidth]{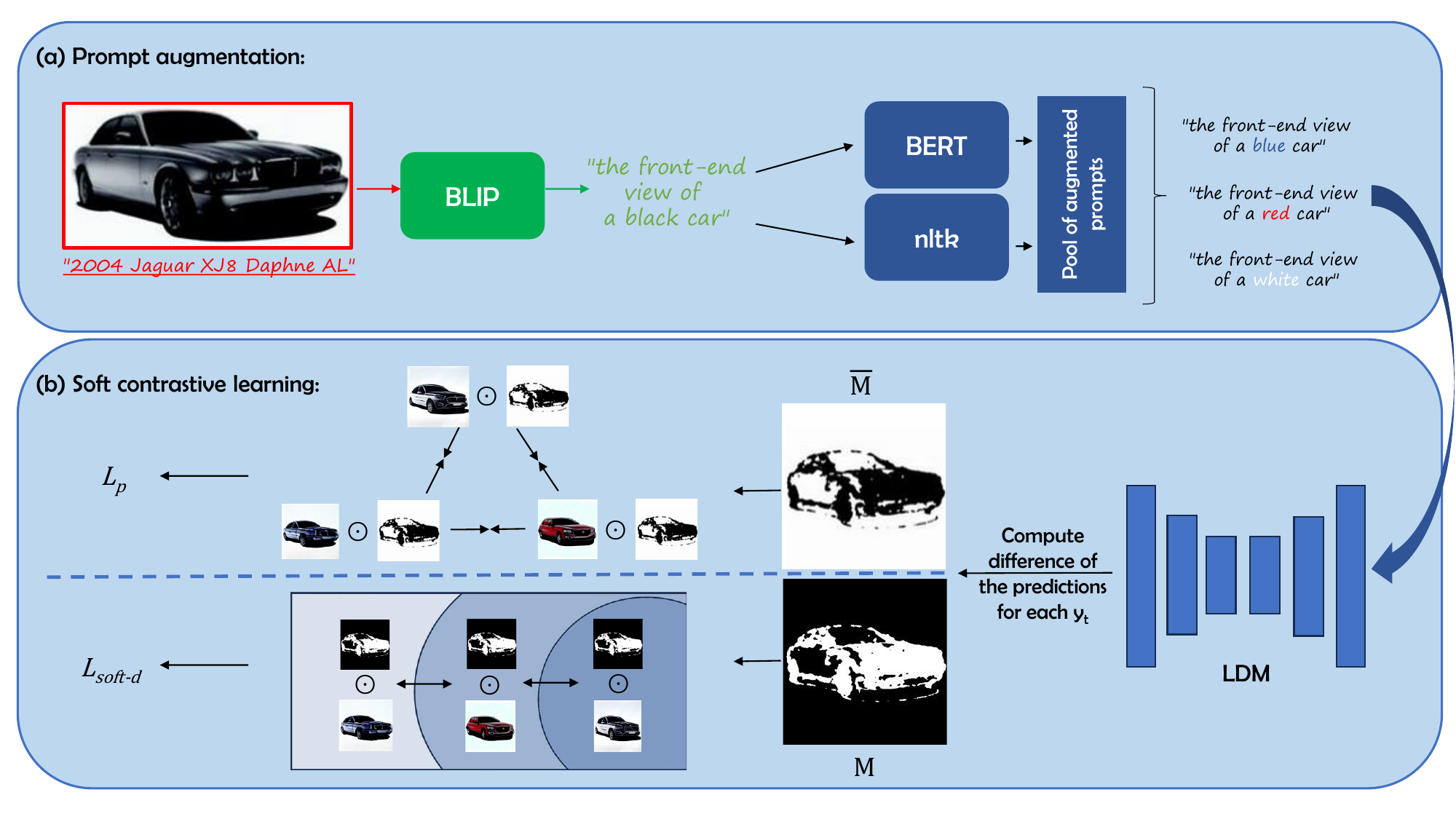}
    \caption{\textbf{Overview of our method.} (a) Prompt Augmentation: In order to augment the prompts to facilitate localised image editing we start by refining textual descriptions for source images using the BLIP captioning model~\cite{li2022blip}, resulting in cleaner captions suitable for further processing. Subsequently, we augment this input prompt by generating a range of target prompts using masked language modeling and exploiting word relations. (b) Soft Contrastive Loss (Soft-CL): The augmented prompts are instrumental in computing an attention mask based on the differences between the generated images. This attention mask is used to bring the inverse masked areas of the generated images closer while pushing away masked areas considering their similarity to the prompts. }
    \label{fig:method_overview}
\end{figure*}

In this section, we present a comprehensive description of our proposed approach. We start by explaining the prompt augmentation process, followed by how these augmented prompts contribute to obtaining an attention mask and training the model using a contrastive loss. We then delve into the finer refinements of our approach, aiming to accommodate the continuous nature of textual prompts through the integration of Soft-CL and soft prompt augmentation. For a visual overview of our methodology, refer to Figure~\ref{fig:method_overview}.

\subsection{Prompt Augmentation}
\label{sec:prompt_augmentation}
We implement prompt augmentation as a critical component of our approach, with the primary goal of generating multiple target prompts to facilitate \emph{self-supervision} for localised image editing. We employ a publicly available large-scale dataset, i.e LAION-5B~\cite{schuhmann2022laionb}, that contains image-caption pairs. However, given the inherent noise in captions from web-derived datasets, manipulating these captions is challenging. To address this, we employ an image captioning model, i.e. BLIP~\cite{li2022blip}, to generate cleaner descriptions associated with the source images. This step is imperative as it enhances the clarity of captions, providing a structured foundation essential for subsequent modifications. For instance, the original caption for the LAION image shown in Figure~\ref{fig:method_overview} is \textit{`2004 Jaguar XJ8 Daphne AL'}, which is not descriptive of the image and cannot be easily manipulated to obtain target prompts. The caption generated with BLIP, \textit{`the front-end view of a black car'}, is cleaner, more descriptive and enables easier manipulation.

With the cleaner captions in hand, we proceed to augment the single input prompt by creating a range of target prompts.  This process begins by masking a random noun or adjective within the input prompt. Leveraging the capabilities of a masked language model, specifically BERT~\cite{devlin2018bert}, we generate potential replacement words for the masked term. Additionally, we enrich our candidate word pool by incorporating semantically related words such as synonyms, antonyms and co-hyponyms, gathered through the NLTK\footnote{https://www.nltk.org/} library. From this pool, we randomly select a set of words to generate multiple variations of the input prompt. For instance, for the aforementioned sample in Figure~\ref{fig:method_overview}, some generated target prompts are: \textit{`the front-end view of a blue car'}, \textit{`the front-end view of a red car'}, \textit{`the front-end view of a white car'}. Please refer to the supplementary material for more samples. These augmented prompts not only diversify the translation capabilities of our network but also lay the groundwork for localised image manipulation by facilitating our self-supervised framework. 
% \vspace{-0.1cm}
\subsection{On-The-Fly Mask Generation}
Diffusion models conditioned on text generate different noise estimates for each prompt. Drawing inspiration from the approach used in DiffEdit during inference, we exploit this characteristic to reveal the intended areas of manipulation within images during training. In our work, we use Latent Diffusion Model (LDM)~\cite{rombach2022sd}, which operates in the latent space to mitigate computational complexities associated with diffusion models. The loss for conditional LDM is given in Eqn.~\ref{eqn:ldm_simple_loss}, where $\epsilon_{\theta}$ represents the noise estimation network, which is a UNet~\cite{ronneberger2015unet}. $\tau_{\theta}$ is a domain-specific encoder projecting the conditioning target prompt to an intermediate representation, and the step $t$ is uniformly sampled from $\{1,2,...,T\}$. The network parameters $\theta$ are optimised to predict the noise ${\epsilon_1}{\sim}\mathcal{N}(0,1)$ that is used for corrupting the encoded version of the input image.
\vspace{-0.3cm}

\begin{equation}
{\mathcal{L}_{LDM}} = \mathbb{E}_{\mathcal{E}(x),y, \epsilon\sim\mathcal{N}(0,1),t}[||\epsilon - \epsilon_{\theta}(z_{t},t,\tau_{\theta}(y))||^2]
\label{eqn:ldm_simple_loss}
\end{equation}

By passing the augmented prompts through the LDM, the model outputs a set of estimated noises. We use the differences between these estimations to delineate the areas of image manipulation. Given the corrupted version of an encoded input image $z_t$ and $N$ augmented target prompts $y_1, y_2, \ldots, y_N$, the noise estimation network, $\epsilon_\theta$, outputs $N$ estimated noises during each iteration. We calculate the average of the differences of these estimation as shown in Eqn.~\ref{eqn:noise_difference} where \(N_p\) represents the number of unique pairs, and it is given by \(N_p = \frac{N \cdot (N-1)}{2}\).
\vspace{-0.3cm}

\begin{equation}
\Delta\epsilon = \frac{1}{N_p} \sum_{i,j} |\epsilon_{\theta}(z_{t},t,\tau_{\theta}(y_i)) - \epsilon_{\theta}(z_{t},t,\tau_{\theta}(y_j))| 
\label{eqn:noise_difference}
\end{equation}

The outcome, $\Delta \epsilon$, provides valuable insights into the varying degrees of noise introduced by different prompts, signifying areas in the image that are most susceptible to modifications. Subsequently, these dissimilarity values are thresholded to establish a binary mask that distinctly highlights regions of alteration within the image. An example can be seen in Figure~\ref{fig:method_overview}, where the mask obtained for the car image outlines only areas that would be manipulated in case the colour of the car would change. In our particular implementation, we employ a threshold empirically set at $0.4$ of the range of absolute differences in noise estimates. The resulting binary mask, $M$, serves as a crucial component for defining the areas targeted for image manipulation within the generated images.

% \begin{equation}
% \Delta\epsilon = | \epsilon_{\theta}(z_{t},t,\tau_{\theta}(y_1)) -  \epsilon_{\theta}(z_{t},t,\tau_{\theta}(y_2))|.
% \end{equation}

\subsection{Contrastive Loss (CL)}
% \vspace{-0.1cm}
Using the mask $M$ derived from augmented prompts, our goal is to guide the model to encourage the masked areas of the latent images, $z_1, z_2, \ldots, z_N$, generated with augmented prompts, $y_1, y_2, \ldots, y_N$, to be in alignment with its respective prompt, while preserving other relevant regions. To achieve this, we introduce a CL that promotes dissimilarity in the masked areas, while drawing the inverse masked areas $\overline{M}$ of the images closer to each other and the source image. The \emph{dissimilarity} within the masked regions is expressed in the first component of the CL as follows:
% By using the mask, $M$, generated through the augmented prompts, our aim is to ensure that the model concentrates on transforming the masked area of the image in accordance with the selected prompt while preserving other pertinent regions. In pursuit of this objective, we introduce a CL, which leverages this mask to encourage manipulations in the designated area by the masks while drawing the inverse masked areas, $\overline{M}$, of the  images $I_1, I_2, \ldots, I_N$ generated using $y_1, y_2, \ldots, y_N$, closer to each other and the source image. Within the CL, the first component encourages \emph{dissimilarity} within the masked regions of the generated images. This part of the loss functions can be expressed as follows:
\vspace{-0.1cm}
\begin{equation}
L_{\text{d}} = \frac{1}{N_p} \sum_{i,j}  \left(1 -\left| (M \odot z_i) - (M \odot z_j) \right|_2\right)
\label{eqn:cl_dissimilarity}
\end{equation}

where $\odot$ represents element-wise multiplication. This term promotes diversity within the areas of the images that should undergo edits, ensuring that the edited content aligns closely with the target prompts. The second component of CL is equally significant. It operates in parallel, aiming to minimise differences within unmasked regions of the images. These regions are critical for \emph{preserving} the parts of the image that should remain unaltered during the editing process. This part of the loss can be defined as:
\vspace{-0.1cm}

\begin{equation}
L_{\text{p}} = \frac{1}{N_p} \sum_{i,j} \left| (\overline{M} \odot z_i) - (\overline{M} \odot z_j) \right|_2 
\end{equation}

The combined CL, unifying both the dissimilarity and preservation components, can be represented as follows:
\vspace{-0.15cm}
\begin{equation}
L_{\text{CL}} =  L_{\text{p}} + \beta \cdot L_{\text{d}} 
\label{eqn:objective_cl}
\end{equation}

The overall objective to fine-tune the LDM is:
\vspace{-0.15cm}
\begin{equation}
L_{\text{obj}} =  L_{LDM} + \alpha \cdot L_{\text{CL}}
\label{eqn:objective}
\end{equation}

\subsection{Soft Contrastive Loss (Soft-CL)}

In image editing, textual prompts present a spectrum of demands that may vary from subtle modifications to profound transformations. For instance, let's consider the difference between transforming `a girl playing in a park' into `a girl playing in a garden' or `a girl playing in a sandbox'. The change from a park to a garden, for example, requires fewer modifications compared to the more extensive transformation into a sandbox. 
To cater to this diversity in editing requirements, we introduce a refinement to our method in the form of Soft Contrastive Loss (Soft-CL). Soft-CL represents an enhancement over the conventional Contrastive Loss, infusing the loss function with the concept of similarity. To accommodate the varying editing demands, we adapt our loss function by revising Eqn.~\ref{eqn:cl_dissimilarity} as follows:
\vspace{-0.15cm}
\begin{equation}
L_{{d}}^{soft} =  \frac{1}{N_p} \sum_{i,j} \left(1 -\left| (M_i \odot z_i) - (M_j \odot z_j)\right|_2^2 \cdot {\gamma}(y_i, y_j)  \right)
\end{equation}

where $\gamma()$ refers to a similarity measure between the prompts, for which we have used the cosine distance between the CLIP embeddings of the prompts. 

In accordance with the Soft-CL we have also enhanced our prompt augmentation strategy to mine prompts in a softer manner rather than selecting them randomly from the pool we construct. For soft prompt augmentation (Soft-PA), we choose the prompts based on differing similarity levels. As we have done in Soft-CL we specifically use the distance between their CLIP embeddings. This process results in a diverse set of prompts, further enhancing the Soft-CL, and consequently the model's ability to handle a wide range of textual inputs to produce contextually relevant image edits.

% \begin{equation}
% \begin{split}
% L = \sum_{i=1}^{N} \sum_{j=1}^{N, j \neq i}&  \left(-\alpha \cdot \left| (M_i \odot I_i) - (M_j \odot I_j) \right|_2^2 \cdot \text{Sim}(i, j) \right.\ \cr
% & \left. + \left|(\mathbf{1} - M_i) \odot I_i - (\mathbf{1} - M_j) \odot I_j\right|_2^2\right)
% \end{split}
% \end{equation}
\vspace{-0.1cm}
\section{Experiments}

\begin{figure*}
  \centering
  \resizebox{0.87\textwidth}{!}{%
    \begin{tabular}{ccccccc}
      input &  & SDEdit & DALL-E 2 & DiffEdit & InstructPix2Pix  & Ours \\
      \includegraphics[width=3cm]{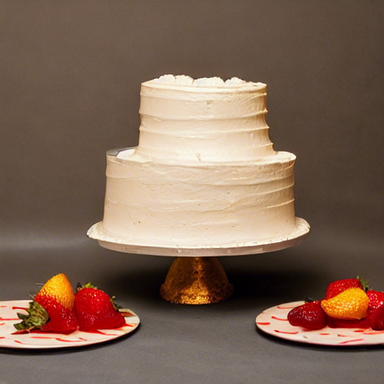} & \raisebox{0.5cm}{\rotatebox{90}{\parbox{2cm}{\centering a \textcolor{red}{strawberry} \\cake}}} & \includegraphics[width=3cm]{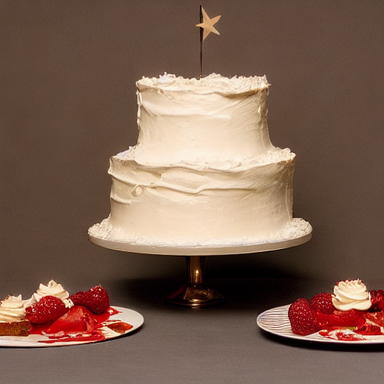} & \includegraphics[width=3cm]{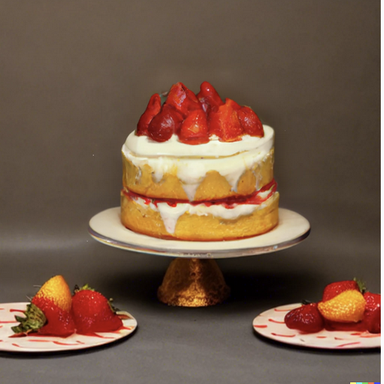} & \includegraphics[width=3cm]{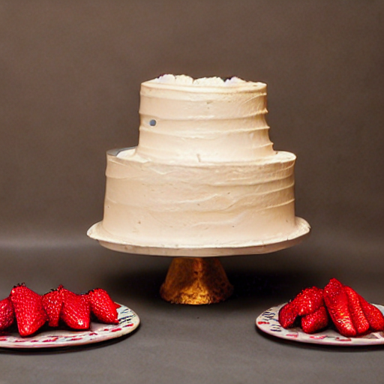} & \includegraphics[width=3cm]{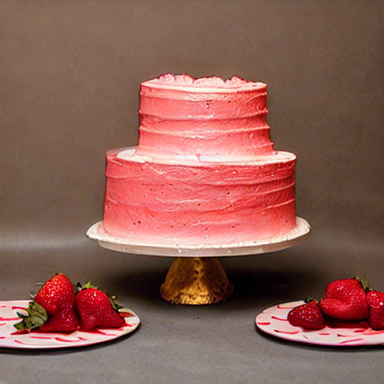} & \includegraphics[width=3cm]{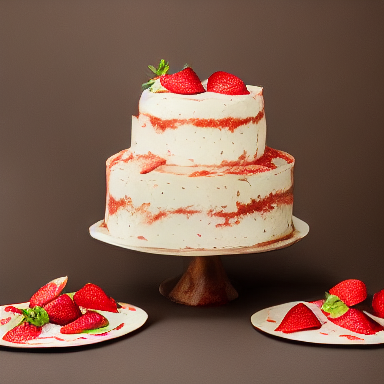} \\
      &  & \includegraphics[width=3cm]{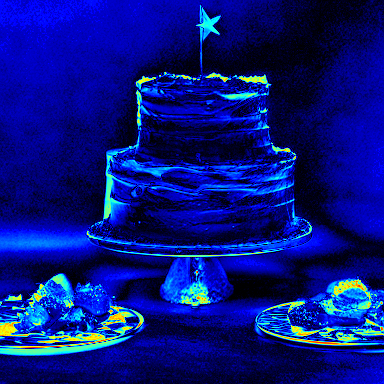} & \includegraphics[width=3cm]{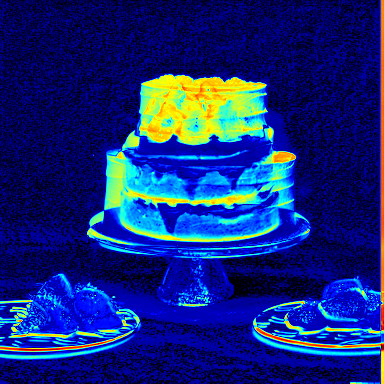} & \includegraphics[width=3cm]{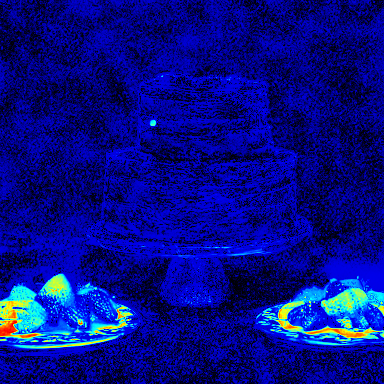} & \includegraphics[width=3cm]{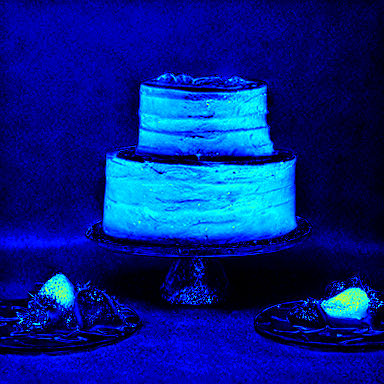} & \includegraphics[width=3cm]{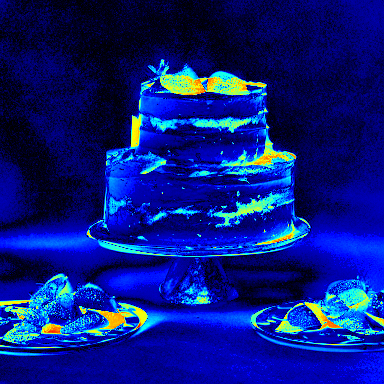} \\
      \includegraphics[width=3cm]{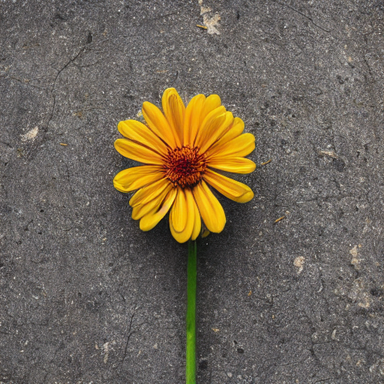} & \raisebox{0.5cm}{\rotatebox{90}{\parbox{2cm}{\centering a yellow \cancel{flower} \textcolor{orange}{rose}}}} & \includegraphics[width=3cm]{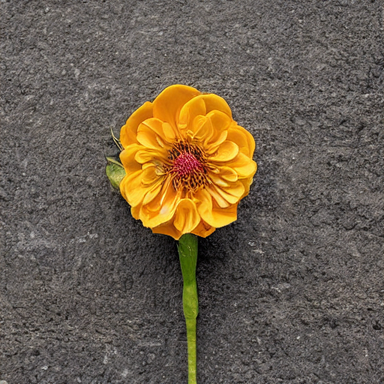} & \includegraphics[width=3cm]{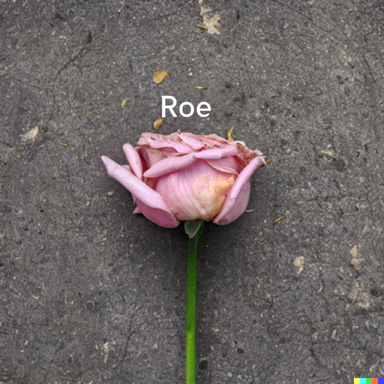} & \includegraphics[width=3cm]{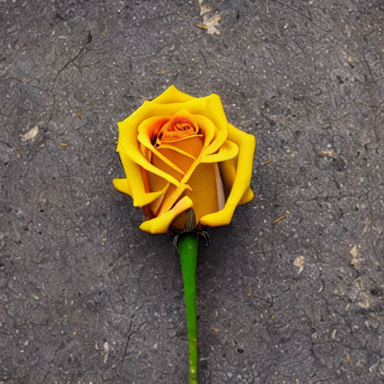} & \includegraphics[width=3cm]{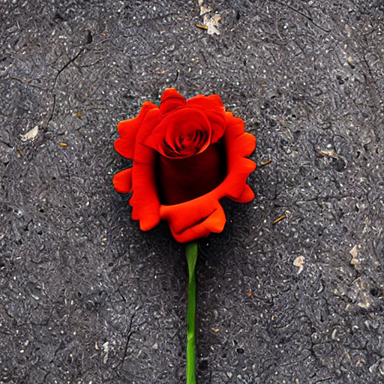} & \includegraphics[width=3cm]{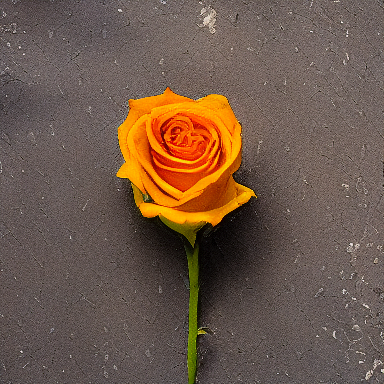}\\
      &  & \includegraphics[width=3cm]{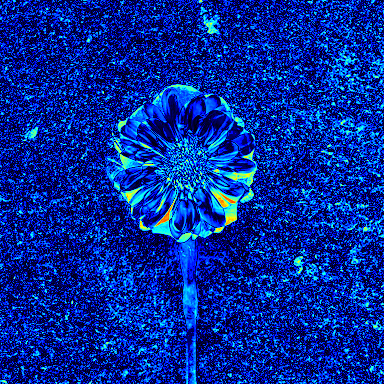} & \includegraphics[width=3cm]{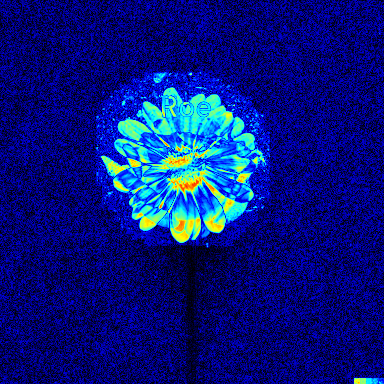} & \includegraphics[width=3cm]{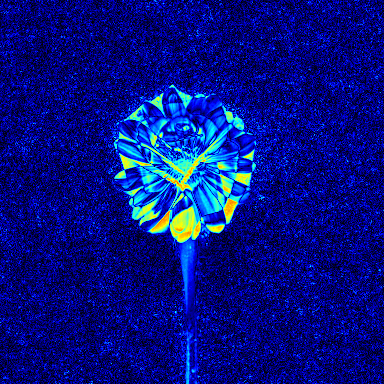} & \includegraphics[width=3cm]{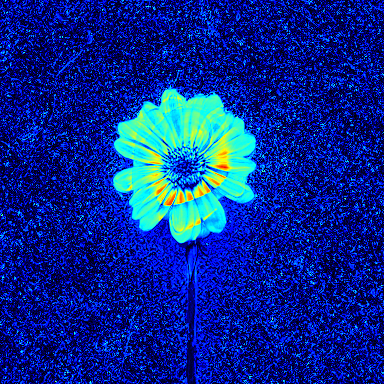} & \includegraphics[width=3cm]{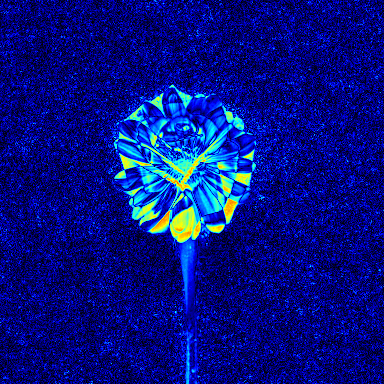} \\
      \includegraphics[width=3cm]{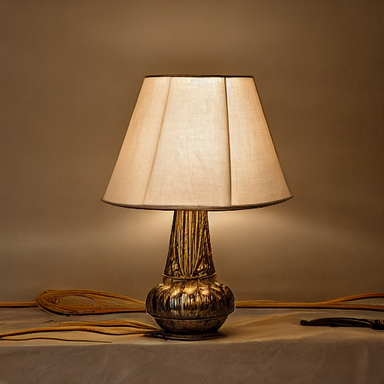} & \raisebox{0.5cm}{\rotatebox{90}{\parbox{2cm}{\centering a \textcolor{brown}{victorian} lamp}}} & \includegraphics[width=3cm]{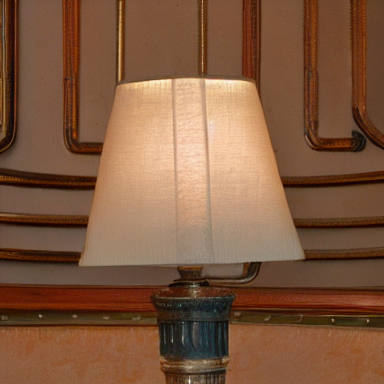} & \includegraphics[width=3cm]{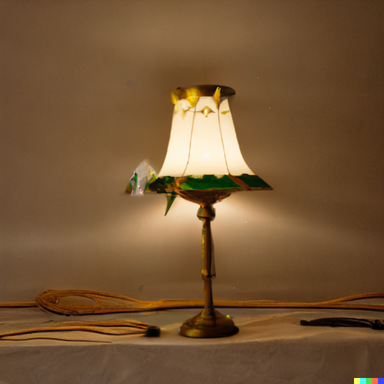} & \includegraphics[width=3cm]{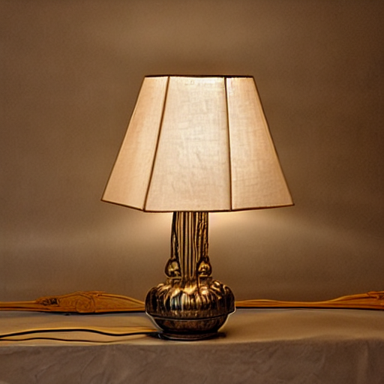} & \includegraphics[width=3cm]{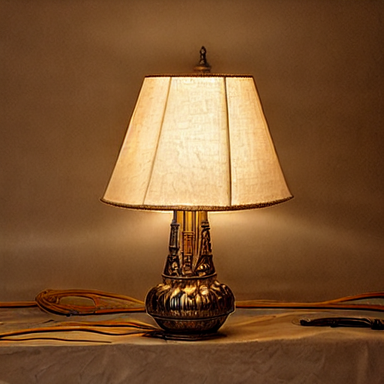} & \includegraphics[width=3cm]{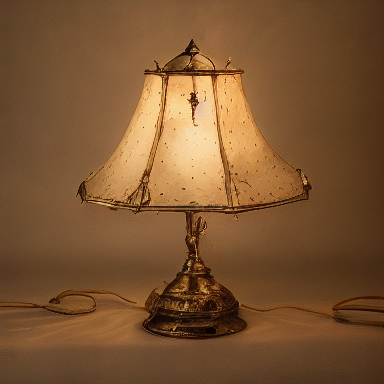}\\      &  & \includegraphics[width=3cm]{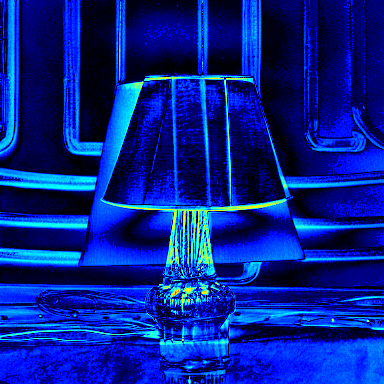} & \includegraphics[width=3cm]{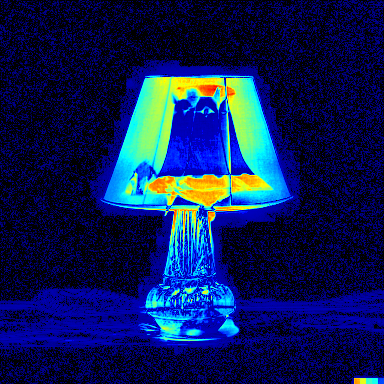} & \includegraphics[width=3cm]{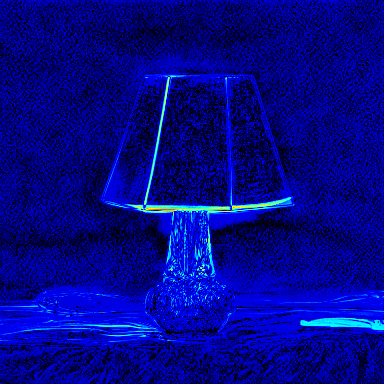} & \includegraphics[width=3cm]{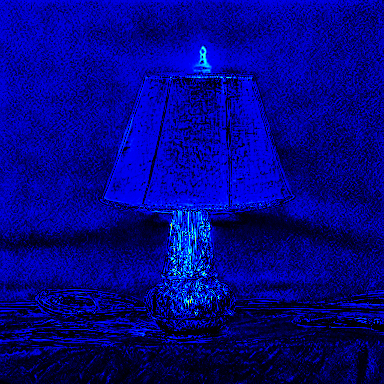} & \includegraphics[width=3cm]{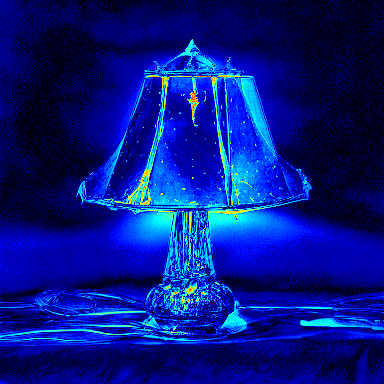} \\
      
      \includegraphics[width=3cm]{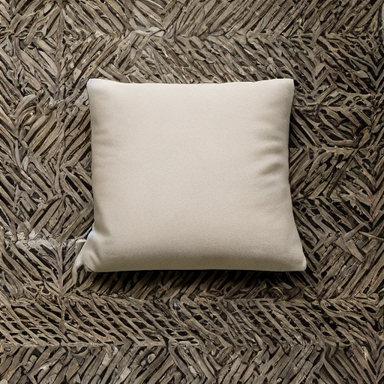} & \raisebox{0.5cm}{\rotatebox{90}{\parbox{2cm}{\centering a \textcolor{green}{green and white check} pillow}}} & \includegraphics[width=3cm]{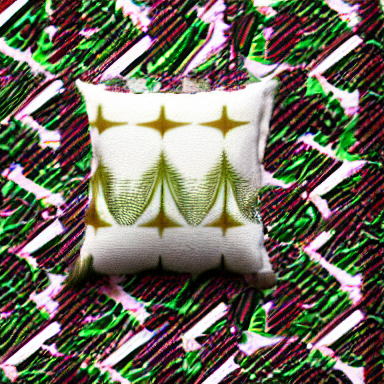} & \includegraphics[width=3cm]{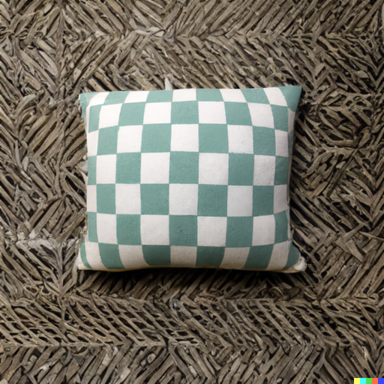} & \includegraphics[width=3cm]{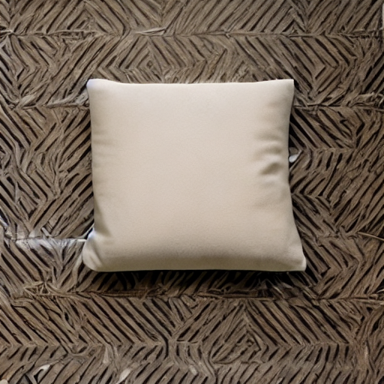} & \includegraphics[width=3cm]{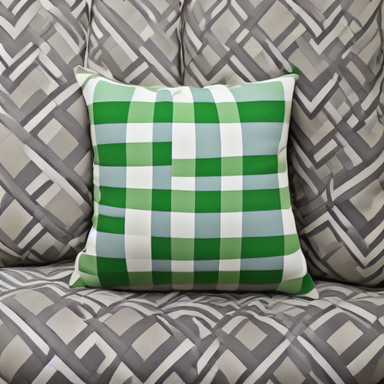} & \includegraphics[width=3cm]{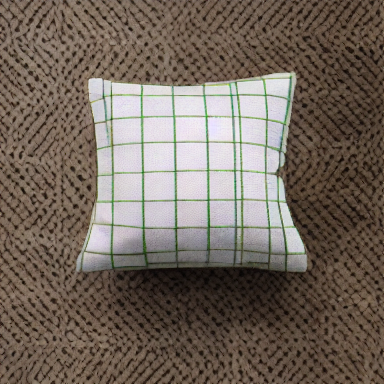}\\      &  & \includegraphics[width=3cm]{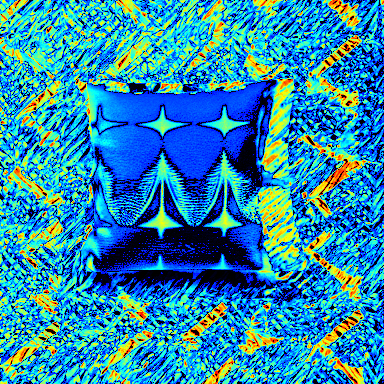} & \includegraphics[width=3cm]{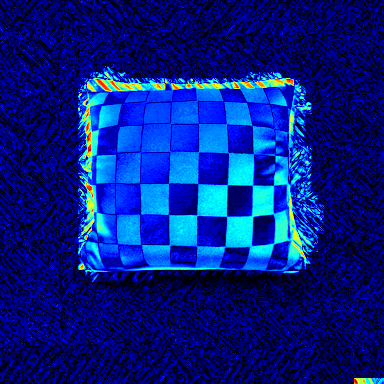} & \includegraphics[width=3cm]{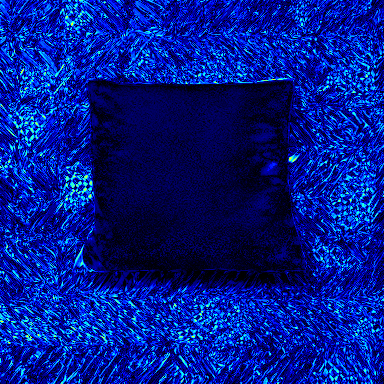} & \includegraphics[width=3cm]{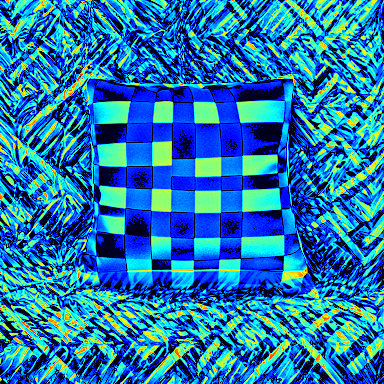} & \includegraphics[width=3cm]{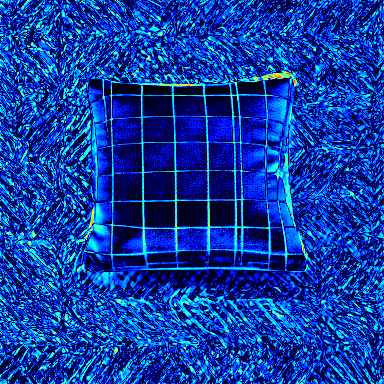} \\
      \includegraphics[width=3cm]{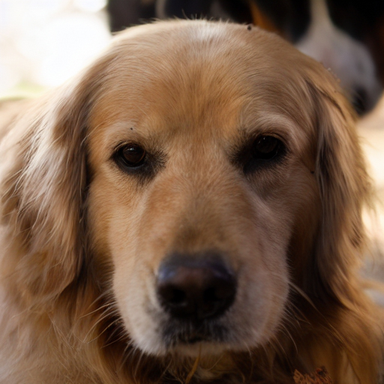} & \raisebox{0.5cm}{\rotatebox{90}{\parbox{2cm}{\centering a \cancel{dog} \textcolor{brown}{bear}}}} & \includegraphics[width=3cm]{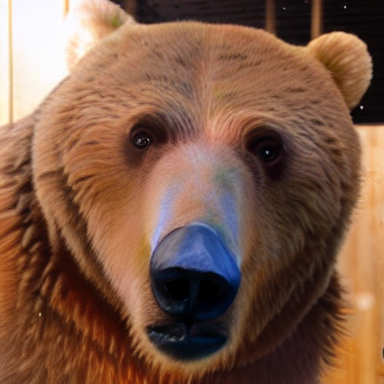} & \includegraphics[width=3cm]{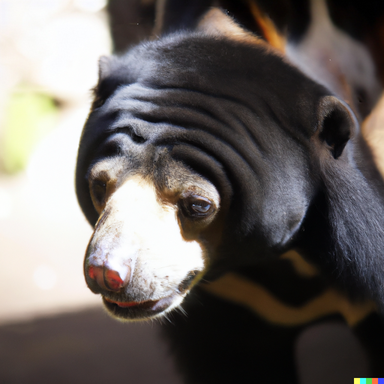} & \includegraphics[width=3cm]{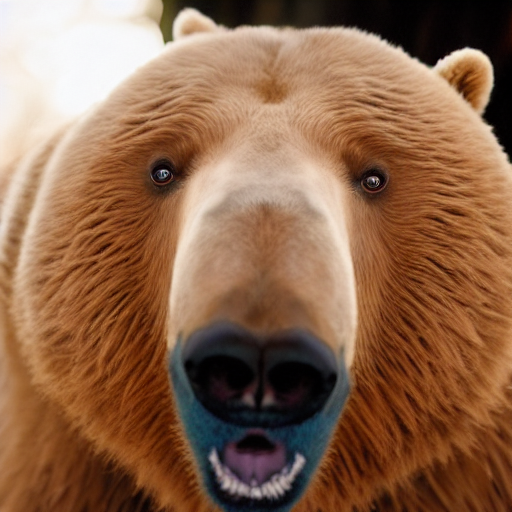} & \includegraphics[width=3cm]{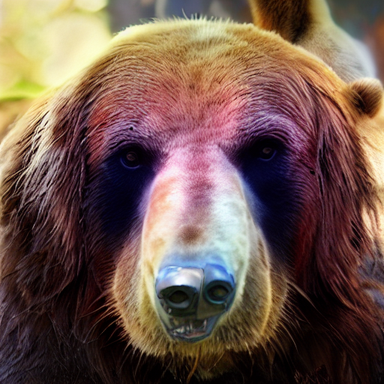} & \includegraphics[width=3cm]{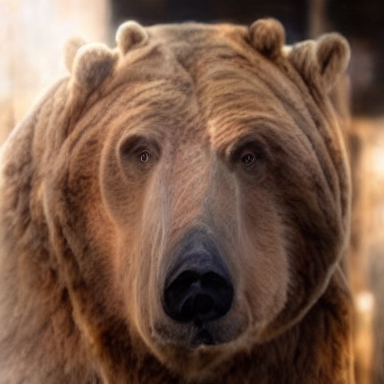}\\      &  & \includegraphics[width=3cm]{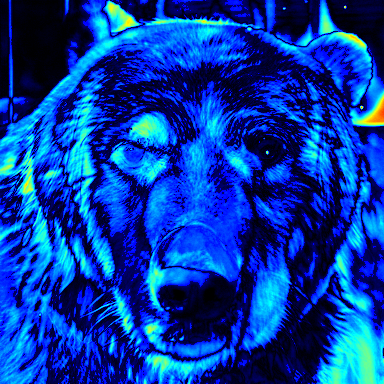} & \includegraphics[width=3cm]{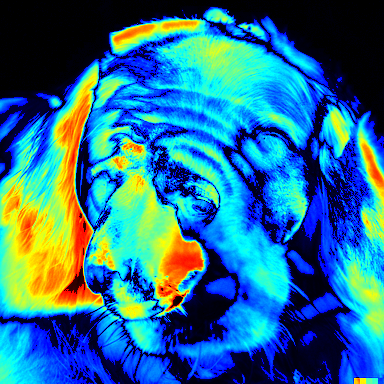} & \includegraphics[width=3cm]{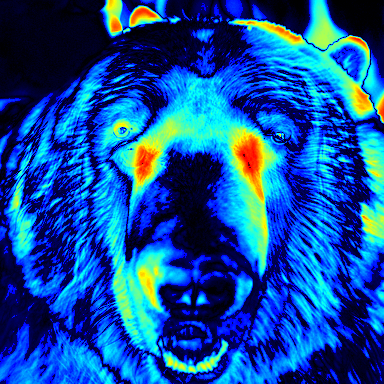} & \includegraphics[width=3cm]{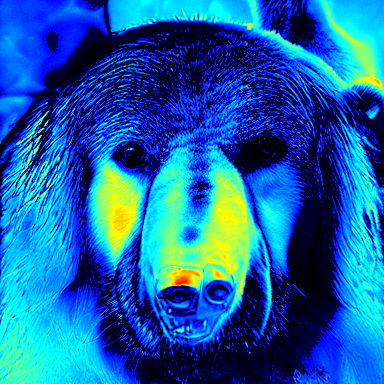} & \includegraphics[width=3cm]{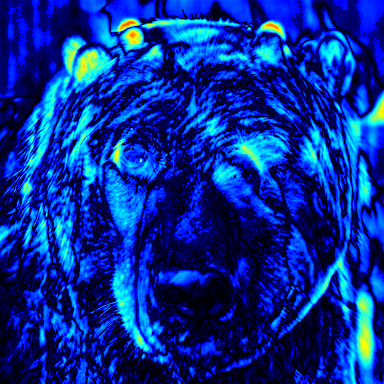} \\
    \end{tabular}%
  }
  \caption{Qualitative comparison of our method against SDEdit~\cite{meng2022sdedit}, DALL-E 2~\cite{ramesh2022dalle}, DiffEdit~\cite{diffedit} and InstructPixtoPix~\cite{brooks2022instructpix2pix} using both generated and real images.}
  \label{fig:qual_sota_comparison}
\end{figure*}

In this section, we analyse our method's performance through experiments, demonstrating its performance with qualitative and quantitative results, comparing to state-of-the-art methods. We also conduct an ablation study, a hyper-parameter study, and discuss limitations.

\vspace{-0.1cm}

\subsection{Experimental Setup}

\noindent\textbf{Datasets.} For training, we used Laion-5B images ~\cite{schuhmann2022laionb} with aesthetics scores greater than 7. For evaluation, we combined images generated using Stable Diffusion v1.4~\cite{rombach2022sd} with real images from COCO~\cite{lin2014coco} and ImageNet~\cite{deng2009imagenet}, totaling 135 images.

\noindent\textbf{Baselines.} Our method is compared against several state-of-the-art baselines, namely SDEdit~\cite{meng2022sdedit}, DiffEdit~\cite{diffedit}, DALL-E 2~\cite{ramesh2022dalle} and InstructPix2Pix~\cite{brooks2022instructpix2pix}. It should be noted that DALL-E 2 and DiffEdit use manually created or automatically generated masks at inference time, and InstructPix2Pix is trained on a paired dataset. Hence, these methods are not directly comparable to ours.

\noindent\textbf{Evaluation Metrics.} To assess the fidelity of our image translation, we utilise CLIPScore~\cite{hessel2021clipscore}, which measures the faithfulness of the image to the target prompt. Additionally, SSIM (Structural Similarity Index) is employed to gauge the faithfulness of the translated image to the input image. We also compute the FID~\cite{seitzer2020fid} in order to evaluate the quality of the generated samples. %In addition to these quantitative metrics, we present qualitative results.% to provide a comprehensive evaluation of our method.

\subsection{Qualitative Results} 
\vspace{-0.1cm}

In Figure~\ref{fig:qual_sota_comparison}, we present a comparison of the results generated by our method against SDEdit~\cite{meng2022sdedit}, DALL-E 2~\cite{ramesh2022dalle}, DiffEdit~\cite{diffedit} and InstructPixtoPix~\cite{brooks2022instructpix2pix} using both generated and real images along wtih heatmaps showing the pixel differences between input an generated images. 
We observe that SDEdit~\cite{meng2022sdedit} often encounters challenges in preserving background details, e.g.~\textit{``a victorian lamp''}. Furthermore, the editing performance of SDEdit may fall short, e.g.~\textit{``a strawberry cake''}. In some scenarios, it struggles with both aspects, as seen in \textit{``a green and white check pillow''}. These limitations can be attributed to the inherent trade-off within SDEdit, balancing between editing and preservation. Additionally, SDEdit is not specifically trained for editing purposes, which contributes to its challenges in maintaining both background integrity and effective editing. 

\begin{table*}
    \centering
    \resizebox{0.76\linewidth}{!}{
    \begin{tabular}{lccccc}
        \toprule
        \textbf{Method} & \textbf{CLIPScore $(\%) \uparrow$} & \textbf{FID $\downarrow$} & \textbf{SSIM-$\overline{M}(\%) \uparrow$} & \textbf{CLIP-R-Precision$(\%) \uparrow$} & \textbf{Human Study}\\
        \midrule
        SDEdit~\cite{meng2022sdedit} & 76.76 & 174 & 63.07 & 75.91 $\pm$ 1.5 & \textcolor{blue}{76.3}\% - \textcolor{red}{23.7}\% \\
        DiffEdit~\cite{diffedit} & 72.77 & 85 & 85.14 & 67.60 $\pm$ 1.7   &\textcolor{blue}{73.84}\% - \textcolor{red}{26.16}\%  \\
        DALL-E 2~\cite{ramesh2022dalle} & 78.47 & 151 & 96.74 & 81.83 $\pm$ 1.4 & \textcolor{blue}{68.3}\% - \textcolor{red}{31.7}\%\\
        InstructPix2Pix~\cite{brooks2022instructpix2pix} & 77.66 & 123 & 81.71 & 77.45 $\pm$ 1.6 & \textcolor{blue}{62.15}\% - \textcolor{red}{37.85}\%  \\
        Our Method & 78.19 & 133 & 70.39 & 80.69 $\pm$ 1.45 & - \\
        \bottomrule
    \end{tabular}}
    \caption{Quantitative comparison of baselines and our method. 
    (↑ higher is better, ↓ lower is better).}
    \label{tab:quantitative-comparison}
    \vspace{-0.5cm}
\end{table*}

As an inpainting method requiring manual user-provided masks for the intended area of manipulation, DALL-E 2~\cite{ramesh2022dalle} falls short in faithfully reproducing the characteristics of the input image, despite its effective preservation of the background. Notably, in instances such as \textit{``a strawberry cake''}, the shape of the cake undergoes significant alterations, compromising the fidelity to the original. Similarly, the orientation of the flower in \textit{``a yellow rose''} is not adequately preserved. These observed shortcomings underscore the challenges inherent in maintaining fidelity to input details in the inpainting-based DALL-E 2 framework. DiffEdit~\cite{diffedit} employs masks derived from input and output prompts during inference. However, DiffEdit's performance proves to be sensitive to the precision of the obtained masks and, consequently, the prompts given by the user. While it yields commendable results when the mask is accurately detected, e.g. \textit{``a yellow rose''}, it faces challenges in other scenarios, such as \textit{``a strawberry cake''}, where the change predominantly occurs in the plates due to inferred mask inaccuracies. It should be noted that DiffEdit, designed primarily for inference time, can be seamlessly integrated into our method. InstructPix2Pix~\cite{brooks2022instructpix2pix} undergoes training with fairly aligned target images generated by P2P~\cite{hertz2022prompt}. Despite its ability to preserve the background and while achieving successful translations in many cases, it carries a notable caveat associated with the use of generated images. The outcomes may exhibit non-realistic qualities, e.g. \textit{``a bear''}, or manipulations can extend beyond the intended region, impacting the entire image, as evident in the check patterned background of \textit{``a green and white check pillow''}. This limitation stems from weaknesses inherited from P2P.%, emphasising the need for careful consideration of its impact on the overall realism of the generated results.
On the other hand, our method demonstrates proficient editing capabilities, preserving the background with minimal undesired changes.
\vspace{-0.1cm}
\subsection{Quantitative Results}
\vspace{-0.1cm}
In Table~\ref{tab:quantitative-comparison}, we present a quantitative comparison of our method with state-of-the-art approaches. The CLIPScore metric indicates the translation coherency regarding the target prompt, while SSIM-$\overline{M}$ scores offer a perspective on background preservation, providing a comprehensive assessment of diverse methodological approaches. SDEdit, serving as our baseline, achieves a CLIPScore of 76.76 and an SSIM-$\overline{M}$ score of 63.07, aligning with qualitative observations of challenges in preserving background details. DiffEdit, relying on automatically generated masks, obtains a CLIPScore of 72.77 and an SSIM-$\overline{M}$ score of 85.14, showing that despite proficiency in background preservation, it falls short in prompt-based translation. DALL-E 2, as an inpainting method with manual masks, demonstrated a CLIPScore of 78.47 and an outstanding SSIM-$\overline{M}$ score of 96.74, showcasing remarkable background preservation due to its inpainting nature. InstructPix2Pix, trained with fairly aligned target images, achieved a CLIPScore of 77.66 and an SSIM-$\overline{M}$ score of 81.71. While competitive, the SSIM-$\overline{M}$ score indicates some compromise in background preservation, aligning with qualitative analysis. Our method, with a CLIPScore of 78.19 and an SSIM-$\overline{M}$ score of 70.39, demonstrates effective editing with a moderate level of background preservation despite not being trained on a paired dataset or using masks at inference. \\
Examining the FID values in Table~\ref{tab:quantitative-comparison} provides valuable insights into the perceptual quality and realism of generated images across different methods. SDEdit reveals a higher FID, indicating a noticeable gap in quality when compared to source images.  DiffEdit, relying on automatically generated masks, presents a lower FID, suggesting a closer resemblance to source images. This lower score may be attributed to its effective use of masks for coherent image generation. DALL-E 2 exhibits a moderate FID, which could be attributed to the challenges of seamlessly blending manipulated regions with the rest of the image, impacting the overall perceptual quality. InstructPix2Pix showcases a competitive FID, maintaining a balance between translation and background preservation. Our method maintains a moderate FID, indicating a balance between effective editing and maintaining a satisfactory level of image quality. \\
We further evaluated our method using CLIP-R-Precision score~\cite{park2021clipprecision} (R=1)(Tab.~\ref{tab:quantitative-comparison}), known for better alignment with human preferences compared to CLIPScore. We observe that it aligns more closely with our qualitative results as our method is outperformed only by DALL-E 2. We argue that, CLIP-Precision-Score still falls short in capturing the identified shortcomings of DALL-E, highlighted in our qualitative findings.\\
Following established protocols~\cite{bar2022text2live, kawar2023imagic}, we conduct a user study on Microworkers, where participants were presented pairs if images and selected the more successful manipulation based on faithfulness to the prompt and input image.
In Tab.~\ref{tab:quantitative-comparison}, \textcolor{blue}{blue} indicates preference for our method and \textcolor{red}{red} for the counterpart baseline. The results align more closely with our qualitative observations than the automatic metrics. DALL-E, despite achieving the highest ClipScore, is preferred less than our method and InstructPix2Pix. This is attributed to its low fidelity to masked content, emphasising nuances that automatic metrics might overlook. Similarly, IP2P's weaknesses, not directly measurable automatically, contribute to its lag behind our method. Further details of this study can be found in the supplementary.

\subsection{Ablation Study}
\vspace{-0.1cm}
\begin{figure}[!t]
  \centering
  \resizebox{\linewidth}{!}{%
    \begin{tabular}{cccccc}
      { Input} && { Baseline} & { $+$ CL} & { $+$ Soft-CL} & { $+$ Soft-PA} \\
      \includegraphics[width=3cm]{figs/qual_res/input/0000.png} & \raisebox{0.5cm}{\rotatebox{90}{\parbox{2cm}{\centering a \textcolor{orange}{orange}\\  cake}}} & \includegraphics[width=3cm]{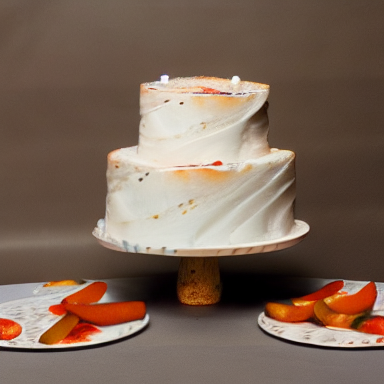} & \includegraphics[width=3cm]{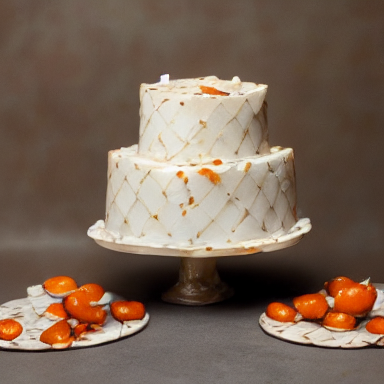} & \includegraphics[width=3cm]{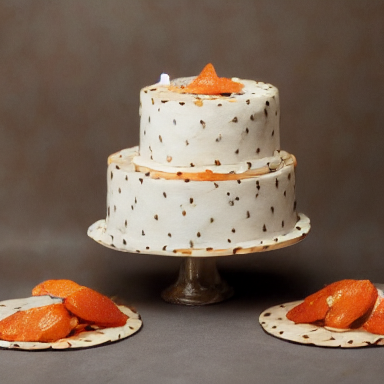}& \includegraphics[width=3cm]{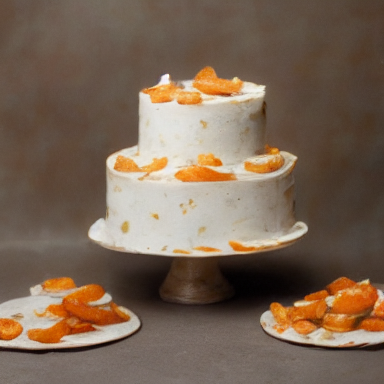}
      \\
      \includegraphics[width=3cm]{figs/qual_res/input/0000.png} & \raisebox{0.5cm}{\rotatebox{90}{\parbox{2cm}{\centering a \textcolor{orange}{mandarin} \\cake}}} & \includegraphics[width=3cm]{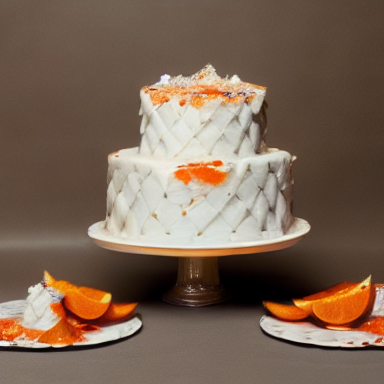} & \includegraphics[width=3cm]{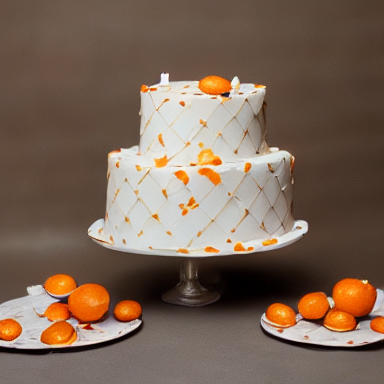} & \includegraphics[width=3cm]{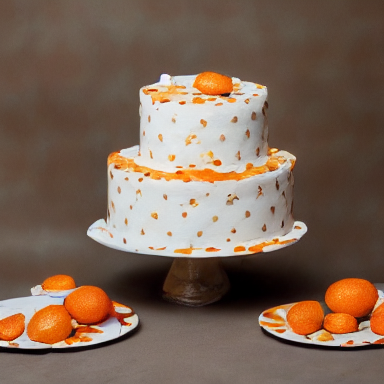} & \includegraphics[width=3cm]{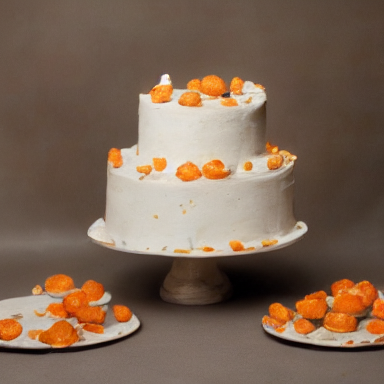}\\
      \includegraphics[width=3cm]{figs/qual_res/input/0000.png} & \raisebox{0.5cm}{\rotatebox{90}{\parbox{2cm}{\centering a \textcolor{brown}{chocolate} \\cake}}} & \includegraphics[width=3cm]{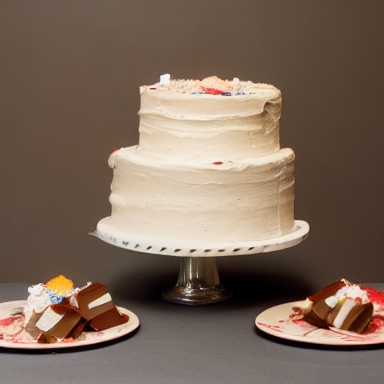} & \includegraphics[width=3cm]{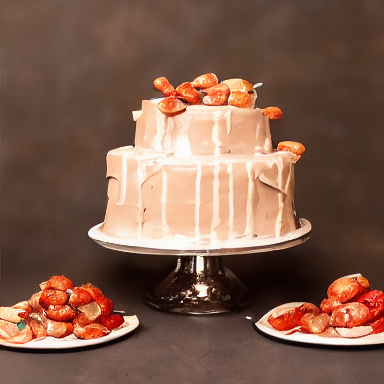} & \includegraphics[width=3cm]{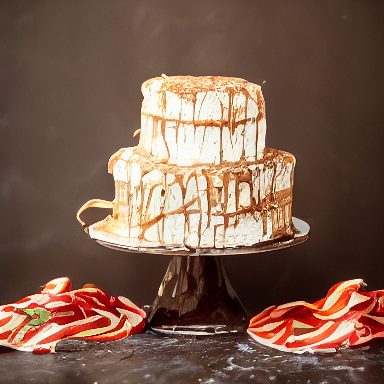} & \includegraphics[width=3cm]{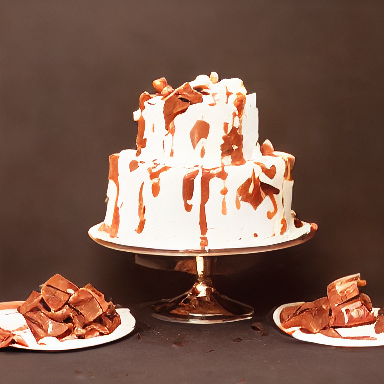}\\
      \includegraphics[width=3cm]{figs/qual_res/input/0000.png} & \raisebox{0.5cm}{\rotatebox{90}{\parbox{2cm}{\centering a \textcolor{brown}{marble} \\cake}}} & \includegraphics[width=3cm]{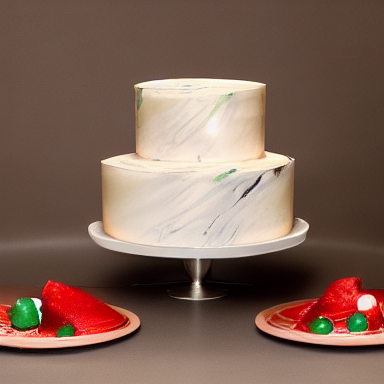} & \includegraphics[width=3cm]{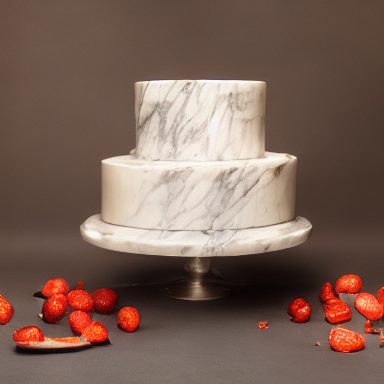} & \includegraphics[width=3cm]{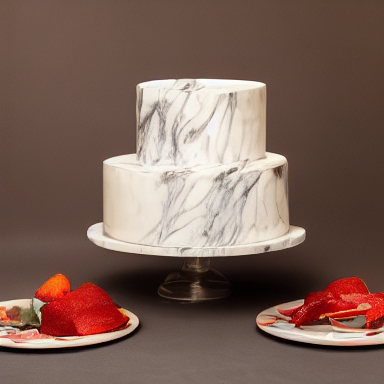} & \includegraphics[width=3cm]{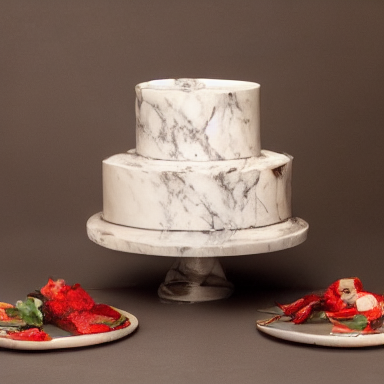}\\

    \end{tabular}%
  }
  \caption{Qualitative comparison of ablation study results.}
  \label{fig:qualitative_ablation}
\end{figure}
Table~\ref{tab:quantitative_ablation} provides insights from our ablation study, investigating the influence of key components in our proposed method. The absence of CL, Soft-CL, Soft-PA is indicated by \xmark{} in the respective columns. Remarkably, the introduction of contrastive loss alone leads to a noticeable enhancement in all metrics, indicating improved translation coherency and content preservation compared to the baseline, which is equivalent to stable diffusion-based SDEdit.While the introduction of soft contrastive loss also shows a rise in all metrics, it becomes superior to CL only with the introduction of soft prompt augmentation, which refines the method's ability to balance translation coherence and background preservation. In summary, our ablation study underscores the pivotal role of contrastive loss in enhancing translation quality. The incorporation of soft contrastive loss, coupled with soft prompt augmentation, contributes to substantial improvements in overall performance.\\
\begin{table}
    \centering
    \resizebox{0.85\linewidth}{!}{
    \begin{tabular}{lcccccc}
        \toprule
        \textbf{CL} & \textbf{Soft-CL} & \textbf{Soft-PA} & \textbf{CLIPScore $(\%) \uparrow$} & \textbf{FID $\downarrow$} & \textbf{SSIM-$\overline{M}(\%) \uparrow$}\\ 
        \midrule
        \xmark & \xmark & \xmark & 76.76 & 174 & 63.07   \\
        \cmark & \xmark & \xmark & 77.94 & 159 &  68.37  \\
        \xmark & \cmark & \xmark & 77.53 & 140 & 67.21   \\
        \xmark & \cmark & \cmark &78.19 & 133 & 70.39   \\
        \bottomrule
    \end{tabular}}
    \caption{Ablation study results for the proposed method components. CL denotes contrastive loss, Soft-CL is soft contrastive loss, and Soft-PA is soft prompt augmentation.}
    \vspace{-0.3 cm}
    \label{tab:quantitative_ablation}
    \vspace{-0.2cm}
\end{table}
In Figure~\ref{fig:qualitative_ablation}, the qualitative ablation study unveils the nuanced impact of each component in refining the translation. The samples, generating different variations of a cake: ``an orange cake'', ``a mandarin cake'', ``a chocolate cake'' and ``a marble cake'' illustrate the model's behavior in the absence of each component. The baseline model produces results with some difference between "orange" and "mandarin," despite their similar semantics. Additionally, it struggles to synthesise a ``chocolate cake'' and ``a marble cake''. The introduction of hard-assigned CL improves generation outputs but does not establish a clear relation between the prompts. With the incorporation of Soft-CL the model becomes adept at discerning fine-grained differences. In this context, ``an orange cake'' and ``a mandarin cake'' demonstrate the model's ability to generate semantically related descriptions, reflecting heightened sensitivity to shared characteristics between these citrus fruits. Soft-PA amplifies these effects by introducing diversity in the input prompts. In the case of ``a chocolate cake'' Soft-PA encourages the model to explore a broader spectrum of possibilities, resulting in varied and contextually rich outputs while showing a clear similarity between `orange' and `mandarin' results. The qualitative examples underscore the role of Soft-CL in enhancing the model's capacity to capture semantic similarities, fostering more coherent and contextually relevant image manipulations.
The interplay between Soft-CL and Soft-PA contributes to the generation of diverse and nuanced outputs, showcasing the effectiveness of these components in refining the editing capabilities of our method.

\subsection{Hyper-parameter Study}

\begin{table}
    \centering
    \resizebox{0.8\linewidth}{!}{
    \begin{tabular}{cccccccc}
        \toprule
        \textbf{Weight ($\alpha$)} & \textbf{CLIPScore $(\%) \uparrow$} & \textbf{FID $\downarrow$} & \textbf{SSIM-$\overline{M}(\%) \uparrow$}\\
        \midrule
        0.25 & 76.11 & 148 & 70.75 \\
        0.5  & 76.54 & 146 & 70.3  \\
        1.0  & 78.20 & 133 & 70.39\\
        2.0  & 72.04 & 121 & 73.22\\
        \bottomrule
    \end{tabular}}
    \caption{Impact of different weight values for $\alpha$ in Eqn.~\ref{eqn:objective} on model performance metrics.}
    \label{tab:hyperparam-results}
    \vspace{-0.6cm}
\end{table}

In Table~\ref{tab:hyperparam-results}, the impact of the weight hyperparameter ($\alpha$ in Eqn.~\ref{eqn:objective}) on our model's performance metrics is shown. Increasing $\alpha$ correlates positively with CLIPScore, indicating improved translation coherency. However, a sharp decrease in CLIPScore is observed with a high value of $\alpha=2.0$. Concurrently, the FID metric decreases with higher $\alpha$ values, reflecting enhanced image quality. The SSIM-$\overline{M}$ score, emphasising fidelity to input images, peaks at $\alpha=2.0$. Our analysis suggests that $\alpha=1.0$ strikes a favorable equilibrium, yielding strong performance across CLIPScore, FID, and SSIM-$\overline{M}$, making it a judicious choice for our method.

\section{Conclusion}
\vspace{-0.1cm}
We introduced a novel method using prompt augmentation for dynamic mask generation and self-supervised learning in image manipulation. Our soft-contrastive loss achieves effective translations in delineated areas while preserving the rest with minimal undesired changes. Addressing challenges in existing methods, our approach provides a promising solution for localised image manipulations. Experimental evaluations, including comparisons with state-of-the-art techniques and ablation studies, were conducted both qualitatively and quantitatively. Our proposed image editing method stands as a significant contribution, paving the way for prompt augmentation and striking a nuanced balance between successful translations and background preservation.\\
\textbf{Acknowledgements.} R. Bodur is funded by the Turkish Ministry of National Education. This work was in part sponsored by NST grant (CRC 21011, MSIT), KOCCA grant (R2022020028, MCST),  IITP grant (RS-2023-00228996, MSIT). BB is supported by NCS University of Aberdeen Startup grant.

% \section{Possible Improvements}

% \subsection{Prompt Augmentation}

% \begin{itemize}
%   \item \textbf{Semantic Similarity}: Currently, prompts are generated by masking a random noun or adjective in the input and replacing it with a related word or a top token from BERT's predictions. To improve this, we can consider incorporating semantic similarity measures between prompts. We can incorporate semantic similarity measures between prompts to specifically serve for the soft CL proposed.
  
%   \item \textbf{Learning the prompt generation process}: We can explore reinforcement learning or other algorithms to fine-tune prompts for generating more relevant textual descriptions for image editing for more coherent and task-specific prompt generation.
% \end{itemize}

% \subsection{Soft Contrastive Loss (Soft CL)}
% \begin{itemize}
%   \item \textbf{Loss Function Variations}: Experiment with different loss functions.
  
%   \subitem \textbf{Multimodal Embeddings}: Utilise multimodal embeddings, like CLIP, to capture richer relationships between images and prompts. A CLIP-based CL.
  
%   \item  \textbf{Semantic Segmentation-Guided Loss}: Incorporate semantic segmentation maps to guide the loss functions. We can ensure that the model's attention aligns with the intended areas of manipulation. 

%   \subitem \textbf{Cross-attention maps:} Using the cross attention maps from the UNet
% \end{itemize}

\newpage
{
    \small
    \bibliographystyle{ieeenat_fullname}
    \bibliography{main}
}


\end{document}